\documentclass{article} 
\usepackage{iclr2016_conference,times}
\usepackage{hyperref}
\usepackage{url}

\usepackage{graphicx}
\usepackage{tabulary}
\usepackage{amsmath}
\usepackage{amsfonts}
\usepackage{epstopdf}

\usepackage{array,multirow,graphicx}

\usepackage{slashbox}
\usepackage{natbib}
\title{Visualizing and Understanding Recurrent Networks}

\author{
Andrej Karpathy\thanks{Both authors contributed equally to this work.} \hspace{0.4in} Justin Johnson\footnotemark[1] \hspace{0.4in} Li Fei-Fei\\
Department of Computer Science, Stanford University\\
\texttt{\small \{karpathy,jcjohns,feifeili\}@cs.stanford.edu}\\
}

\begin{document}

\maketitle

\begin{abstract}
Recurrent Neural Networks (RNNs), and specifically a variant with Long Short-Term Memory (LSTM),
are enjoying renewed interest as a result of
successful applications in a wide range of machine learning problems that involve sequential data. However,
while LSTMs provide exceptional results in practice, the source of their performance and their limitations
remain rather poorly understood. Using character-level language models as an interpretable testbed,
we aim to bridge this gap by providing an analysis of their representations, predictions and error types.
In particular, our experiments reveal the existence of interpretable cells that keep track of long-range
dependencies such as line lengths, quotes and brackets. Moreover, our comparative analysis with finite horizon
$n$-gram models traces the source of the LSTM improvements to long-range structural dependencies.
Finally, we provide analysis of the remaining errors and suggests areas for further study.
\end{abstract}

\section{Introduction}

Recurrent Neural Networks, and specifically a variant with Long Short-Term Memory (LSTM) \cite{lstm},
have recently emerged as an effective model in a wide variety of applications that involve sequential data.
These include language modeling \cite{mikolov}, handwriting recognition and generation \cite{graves},
machine translation \cite{sutskever2014sequence,bahdanau2014neural}, speech recognition \cite{graves2013speech},
video analysis \cite{donahue2014long} and image captioning \cite{vinyals2014show,karpathy2014deep}.

However, both the source of their impressive performance and their shortcomings remain poorly understood. This raises
concerns of the lack of interpretability and limits our ability design better architectures. A few recent ablation studies
analyzed the effects on performance as various gates and connections are removed \cite{odyssey,chung2014empirical}.
However, while this analysis illuminates the performance-critical pieces of the architecture, it is still limited to
examining the effects only on the global level of the final test set perplexity alone.
Similarly, an often cited advantage of the LSTM architecture is that it can store and retrieve information over long
time scales using its gating mechanisms, and this ability has been carefully studied in toy settings \cite{lstm}.
However, it is not immediately clear that similar mechanisms can be effectively discovered and utilized
by these networks in real-world data, and with the common use of simple stochastic gradient descent and
truncated backpropagation through time.

To our knowledge, our work provides the first empirical exploration
of the predictions of LSTMs and their learned representations on real-world data. Concretely, we use character-level
language models as an interpretable testbed for illuminating the long-range dependencies learned by LSTMs.
Our analysis reveals the existence of cells that robustly identify interpretable, high-level patterns such as
line lengths, brackets and quotes. We further quantify the LSTM predictions with comprehensive comparison
to $n$-gram models, where we find that LSTMs perform significantly better on characters that require long-range
reasoning. Finally, we conduct an error analysis in which we \textit{``peel the onion''} of errors with a
sequence of oracles. These results allow us to quantify the extent of remaining errors in several categories and
to suggest specific areas for further study.

\vspace{-0.1in}
\section{Related Work}
\vspace{-0.1in}

\textbf{Recurrent Networks}. Recurrent Neural Networks (RNNs) have a long history of applications in various
sequence learning tasks \cite{rnn,dlbook,rumelhart1985learning}. Despite their early successes, the
difficulty of training simple recurrent networks \cite{bengiornn94,pascanu2012difficulty} has encouraged various proposals for improvements
to their basic architecture. Among the most successful variants are the Long Short Term Memory networks \cite{lstm}, which
can in principle store and retrieve information over long time periods with explicit gating mechanisms and a
built-in constant error carousel. In the recent years there has been a renewed interest in further improving on the basic architecture by
modifying the functional form as seen with Gated Recurrent Units \cite{gru}, incorporating content-based soft attention
mechanisms \cite{bahdanau2014neural,memorynets}, push-pop stacks \cite{armand}, or more generally external memory arrays
with both content-based and relative addressing mechanisms \cite{ntm}. In this work we focus the majority of our analysis on the
LSTM due to its widespread popularity and a proven track record.

\textbf{Understanding Recurrent Networks}. While there is an abundance of work that modifies or extends the basic LSTM
architecture, relatively little attention has been paid to understanding the properties of its representations and predictions.
\cite{odyssey} recently conducted a comprehensive study of LSTM components. Chung et al. evaluated GRU
compared to LSTMs \cite{chung2014empirical}. \cite{jozefowicz2015empirical} conduct an automated architecture search of thousands of RNN architectures.
\cite{rnndepth} examined the effects of depth . These approaches study recurrent network based only on the variations
in the final test set cross entropy, while we break down the performance into interpretable categories and study individual error types.
Most related to our work is \cite{hermans2013training}, who also study the long-term interactions learned by recurrent networks in
the context of character-level language models, specifically in the context of parenthesis closing and time-scales analysis. Our work complements
their results and provides additional types of analysis. Lastly, we are heavily influenced by work on in-depth analysis of errors in object detection
\cite{hoiem2012diagnosing}, where the final mean average precision is similarly broken down and studied in detail.

\vspace{-0.1in}
\section{Experimental Setup}
\vspace{-0.1in}

We first describe three commonly used recurrent network architectures (RNN, LSTM and the GRU),
then describe their used in sequence learning and finally discuss the optimization.

\vspace{-0.1in}
\subsection{Recurrent Neural Network Models}
\vspace{-0.15in}

The simplest instantiation of a deep recurrent network arranges hidden state vectors $h_t^l$ in a two-dimensional grid,
where $t = 1 \ldots T$ is thought of as time and $l = 1 \ldots L$ is the depth. The bottom row of vectors $h_t^0 = x_t$
at depth zero holds the input vectors $x_t$ and each vector in the top row $\{h_t^L\}$ is used to predict an
output vector $y_t$. All intermediate vectors $h_t^l$ are computed with a recurrence formula based on $h_{t-1}^l$ and $h_t^{l-1}$.
Through these hidden vectors, each output $y_t$ at time step $t$ becomes a function of all input vectors up to $t$,
$\{x_1, \ldots, x_t \}$. The precise mathematical form of the recurrence $(h_{t-1}^l$ , $h_t^{l-1}) \rightarrow h_t^l$ varies from
model to model and we describe these details next.

\textbf{Vanilla Recurrent Neural Network} (RNN) has a recurrence of the form

\vspace{-0.15in}
\begin{align*}
h^l_t = \tanh W^l \begin{pmatrix}h^{l - 1}_t\\h^l_{t-1}\end{pmatrix}
\end{align*}
\vspace{-0.15in}

where we assume that all $h \in \mathbb{R}^n$. The parameter matrix $W^l$ on each layer has dimensions [$n \times 2n$] and $\tanh$
is applied elementwise. Note that $W^l$ varies between layers but is shared through time. We omit the bias vectors for brevity.
Interpreting the equation above, the inputs from the layer
below in depth ($h_t^{l-1}$) and before in time ($h_{t-1}^l$) are transformed and interact through additive interaction before being squashed
by $\tanh$. This is known to be a weak form of coupling~\cite{mrnn}. Both the LSTM and the GRU (discussed next) include
more powerful multiplicative interactions.

\textbf{Long Short-Term Memory} (LSTM)~\cite{lstm} was designed to address the difficulties of training RNNs~\cite{bengiornn94}.
In particular, it was observed that the backpropagation dynamics caused the gradients in an RNN to either vanish or explode.
It was later found that the exploding gradient concern can be alleviated with a heuristic of clipping the gradients at some maximum value \cite{pascanu2012difficulty}.
On the other hand, LSTMs were designed to mitigate the vanishing gradient problem. In addition to a hidden state vector $h_t^l$, LSTMs also maintain
a memory vector $c_t^l$. At each time step the LSTM can choose to read from, write to, or reset the cell using explicit
gating mechanisms. The precise form of the update is as follows:

\vspace{-0.1in}
\begin{minipage}{.5\linewidth}
\begin{align*}
&\begin{pmatrix}i\\f\\o\\g\end{pmatrix} =
\begin{pmatrix}\mathrm{sigm}\\\mathrm{sigm}\\\mathrm{sigm}\\\tanh\end{pmatrix}
W^l \begin{pmatrix}h^{l - 1}_t\\h^l_{t-1}\end{pmatrix}
\end{align*}
\end{minipage}%
\begin{minipage}{.5\linewidth}
\begin{align*}
&c^l_t = f \odot c^l_{t-1} + i \odot g\\
&h^l_t = o \odot \tanh(c^l_t)
\end{align*}
\end{minipage}
\vspace{-0.1in}

Here, the sigmoid function $\mathrm{sigm}$ and $\tanh$ are applied element-wise, and $W^l$ is a [$4n \times 2n$] matrix.
The three vectors $i,f,o \in \mathbb{R}^n$ are thought of as binary gates that control whether each memory cell is updated,
whether it is reset to zero, and whether its local state is revealed in the hidden vector, respectively. The activations of these gates
are based on the sigmoid function and hence allowed to range smoothly between zero and one to keep the model differentiable.
The vector $g \in \mathbb{R}^n$ ranges between -1 and 1 and is used to additively modify the memory contents. This additive interaction is a critical feature
of the LSTM's design, because during backpropagation a sum operation merely distributes gradients. This allows gradients on the
memory cells $c$ to flow backwards through time uninterrupted for long time periods, or at least until the flow is disrupted with the
multiplicative interaction of an active forget gate. Lastly, note that an implementation of the LSTM requires one to maintain
two vectors ($h_t^l$ and $c_t^l$) at every point in the network.

\textbf{Gated Recurrent Unit} (GRU) \cite{gru} recently proposed as a simpler alternative to the LSTM that takes the form:

\vspace{-0.1in}
\begin{minipage}{.5\linewidth}
\begin{align*}
&\begin{pmatrix}r\\z\end{pmatrix} =
\begin{pmatrix}\mathrm{sigm}\\\mathrm{sigm}\end{pmatrix}
W_r^l \begin{pmatrix}h^{l - 1}_t\\h^l_{t-1}\end{pmatrix}
\end{align*}
\end{minipage}%
\begin{minipage}{.5\linewidth}
\begin{align*}
&\tilde{h}^l_t = \tanh( W_x^l h^{l - 1}_t+ W_g^l ( r \odot h^l_{t-1}) ) \\
& h^l_t = (1 - z) \odot h^l_{t-1} + z \odot \tilde{h}^l_t
\end{align*}
\end{minipage}
\vspace{-0.1in}

Here, $W_r^l$ are [$2n \times 2n$] and $W_g^l$ and $W_x^l$ are [$n \times n$]. The GRU has the interpretation of
computing a \textit{candidate} hidden vector $\tilde{h}^l_t$ and then smoothly interpolating towards it gated by $z$.

\vspace{-0.1in}
\subsection{Character-level Language Modeling}
\vspace{-0.1in}

We use character-level language modeling as an interpretable testbed for sequence learning. In this setting, the input to
the network is a sequence of characters and the network is trained to predict the next character in the sequence with
a Softmax classifier at each time step. Concretely, assuming a fixed vocabulary of $K$ characters we encode all characters
with $K$-dimensional 1-of-$K$ vectors $\{x_t\}, t = 1, \ldots, T$, and feed these to the recurrent network to obtain
a sequence of $D$-dimensional hidden vectors at the last layer of the network $\{h_t^L\}, t = 1, \dots, T$. To obtain
predictions for the next character in the sequence we project this top layer of activations to a sequence of vectors $\{y_t\}$, where
$y_t = W_y h_t^L$ and $W_y$ is a [$K \times D$] parameter matrix. These vectors are interpreted as holding the (unnormalized)
log probability of the next character in the sequence and the objective is to minimize the average cross-entropy loss over all targets.

\vspace{-0.1in}
\subsection{Optimization}
\label{subsec:optimization}
\vspace{-0.1in}

Following previous work \cite{sutskever2014sequence} we initialize all parameters uniformly in range $[-0.08, 0.08]$. We use mini-batch
stochastic gradient descent with batch size 100 and RMSProp \cite{rmsprop} per-parameter adaptive
update with base learning rate $2 \times 10 ^{-3}$ and decay $0.95$. These settings work robustly with all of our models.
The network is unrolled for 100 time steps. We train each model for 50 epochs and decay the learning rate after 10 epochs by multiplying it with a factor of
0.95 after each additional epoch. We use early stopping based on validation performance and cross-validate the amount of
dropout for each model individually.

\vspace{-0.1in}
\section{Experiments}
\vspace{-0.1in}

\textbf{Datasets}.
Two datasets previously used in the context of character-level language models are the Penn Treebank
dataset \cite{marcus1993building} and the Hutter Prize 100MB of Wikipedia dataset \cite{hutter} . However,
both datasets contain a mix of common language and special markup. Our goal is not to compete with previous
work but rather to study recurrent networks in a controlled setting and on both ends on the
spectrum of degree of structure. Therefore, we chose to use Leo Tolstoy's \textit{War and Peace} (WP) novel,
which consists of 3,258,246 characters of almost entirely English text with minimal markup, and at the other
end of the spectrum the source code of the \textit{Linux Kernel} (LK).
We shuffled all header and source files randomly and concatenated them into a single file to form
the 6,206,996 character long dataset. We split the data into train/val/test splits as 80/10/10 for WP and
90/5/5 for LK. Therefore, there are approximately 300,000 characters in the validation/test splits in each case. The
total number of characters in the vocabulary is 87 for WP and 101 for LK.

\vspace{-0.1in}
\subsection{Comparing Recurrent Networks}
\vspace{-0.1in}

We first train several recurrent network models to support further analysis and to compare their performance
in a controlled setting. In particular, we train models in the cross product of type (LSTM/RNN/GRU), number of layers
(1/2/3), number of parameters (4 settings), and both datasets (WP/KL). For a 1-layer LSTM we used hidden size
vectors of 64,128,256, and 512 cells, which with our character vocabulary sizes translates to approximately
50K, 130K, 400K, and 1.3M parameters respectively. The sizes of hidden layers of the other models were carefully
chosen so that the total number of parameters in each case is as close as possible to these 4 settings.

\begin{figure*}[t]
\centering
\begin{minipage}{.7\textwidth}
\centering
\small
\begin{tabulary}{\linewidth}{LCCC|CCC|CCC}
& \multicolumn{3}{c}{\textbf{LSTM}} & \multicolumn{3}{c}{\textbf{RNN}} & \multicolumn{3}{c}{\textbf{GRU}} \\
Layers & 1 & 2 & 3 & 1 & 2 & 3 & 1 & 2 & 3 \\
\hline
\hline
Size & \multicolumn{9}{c}{War and Peace Dataset} \\
\hline
64 & 1.449 & 1.442 & 1.540 & 1.446 & 1.401 & 1.396 & 1.398 & \textbf{1.373} & 1.472 \\
128 & 1.277 & 1.227 & 1.279 & 1.417 & 1.286 & 1.277 & 1.230 & \textbf{1.226} & 1.253 \\
256 & 1.189 & \textbf{1.137} & 1.141 & 1.342 & 1.256 & 1.239 & 1.198 & 1.164 & 1.138 \\
512 & 1.161 & 1.092 & 1.082 & - & - & - & 1.170 & 1.201 & \textbf{1.077} \\
\hline
\multicolumn{10}{c}{Linux Kernel Dataset} \\
\hline
64 & 1.355 & \textbf{1.331} & 1.366 & 1.407 & 1.371 & 1.383 & 1.335 & 1.298 & 1.357 \\
128 & 1.149 & 1.128 & 1.177 & 1.241 & \textbf{1.120} & 1.220 & 1.154 & 1.125 & 1.150 \\
256 & 1.026 & \textbf{0.972} & 0.998 & 1.171 & 1.116 & 1.116 & 1.039 & 0.991 & 1.026 \\
512 & 0.952 & 0.840 & 0.846 & - & - & - & 0.943 & 0.861 & \textbf{0.829} \\
\end{tabulary}
\end{minipage}%
\begin{minipage}{0.3\textwidth}
\centering
\vspace{0.24in}
\includegraphics[width=1\textwidth]{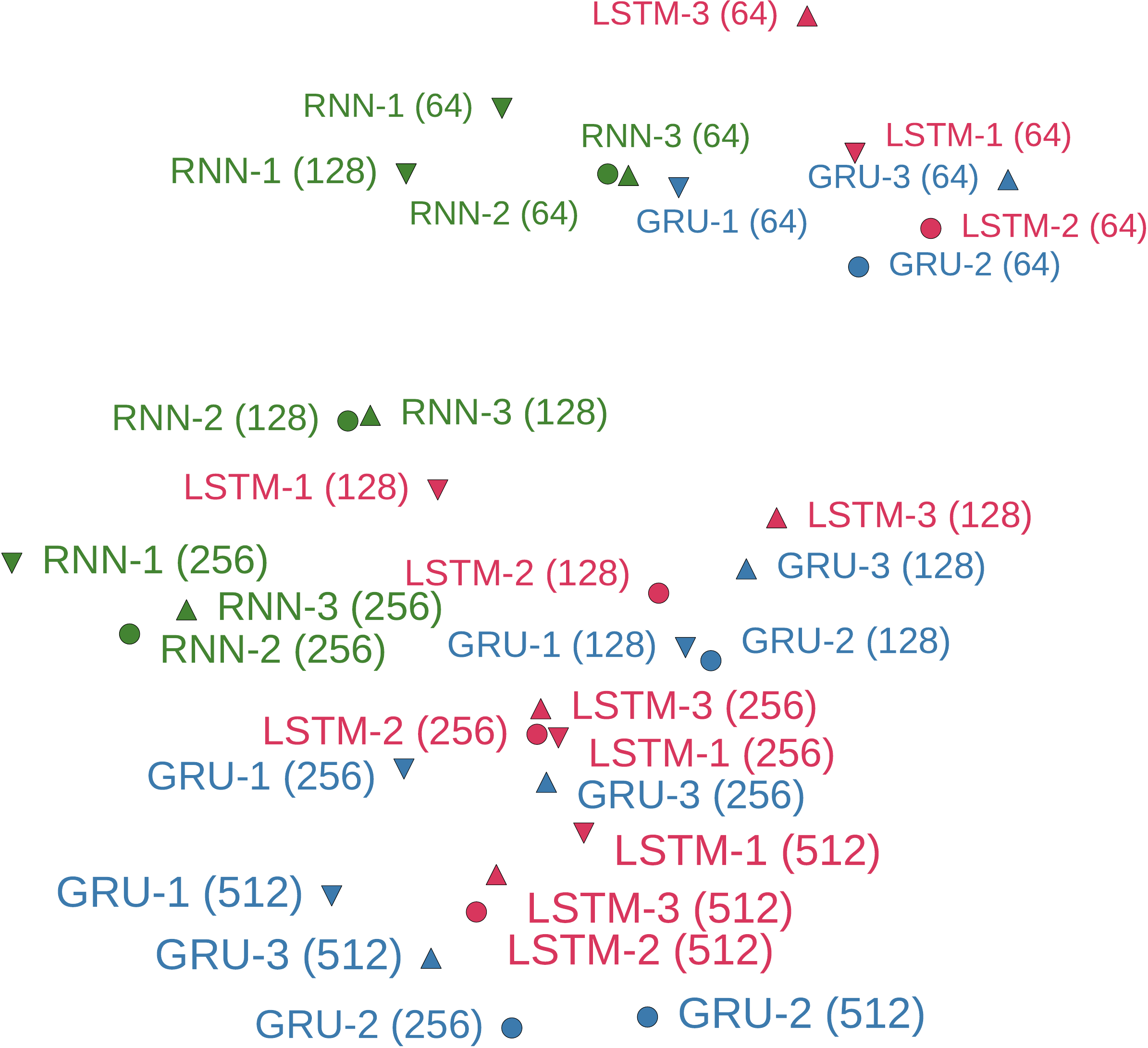}
\end{minipage}
\caption{\textbf{Left:} The \textbf{test set cross-entropy loss} for all models and datasets (low is good).
Models in each row have nearly equal number of parameters. The test set has 300,000 characters.
The standard deviation, estimated with 100 bootstrap samples, is less than $4\times10^{-3}$ in all cases.
\textbf{Right:} A t-SNE embedding based on the probabilities assigned to
test set characters by each model on War and Peace. The color, size, and marker correspond to model type,
model size, and number of layers.}
\label{fig:performance}
\vspace{-0.2in}
\end{figure*}

The test set results are shown in Figure \ref{fig:performance}.
Our consistent finding is that depth of at least two is beneficial. However, between two and three layers our
results are mixed. Additionally, the results are mixed between the LSTM and the GRU, but both significantly
outperform the RNN. We also computed the fraction of times that each pair of models
agree on the most likely character and use it to render a t-SNE \cite{tsne} embedding (we found this more
stable and robust than the KL divergence). The plot (Figure \ref{fig:performance}, right) further supports
the claim that the LSTM and the GRU make similar predictions while the RNNs form their own cluster.

\renewcommand\tabcolsep{3pt} 

\renewcommand\tabcolsep{6pt} 

\vspace{-0.1in}
\subsection{Internal Mechanisms of an LSTM}
\vspace{-0.1in}

\begin{figure*}
\includegraphics[width=1\linewidth]{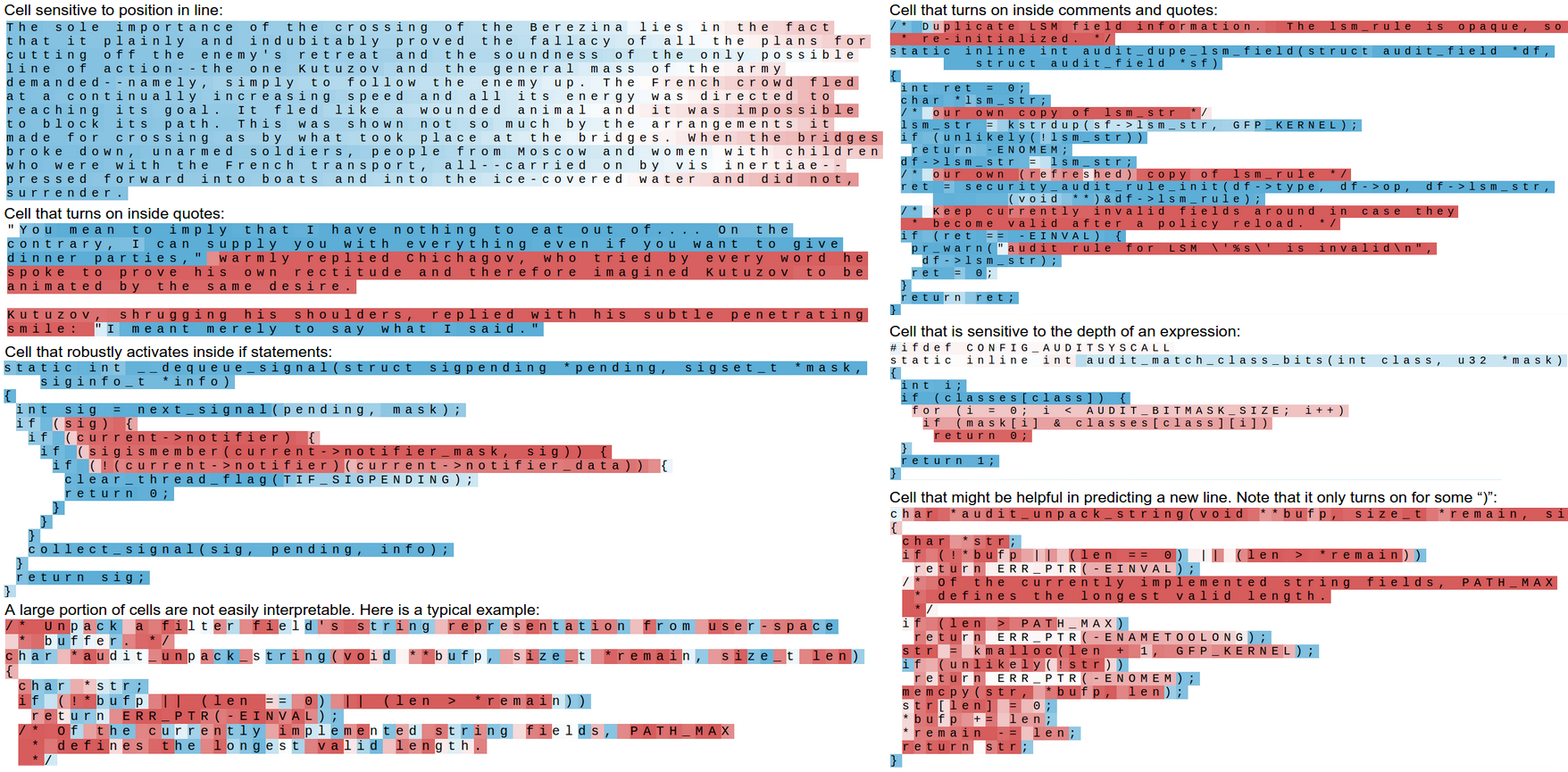}
\caption{Several examples of cells with interpretable activations discovered in our best Linux Kernel and
War and Peace LSTMs. Text color corresponds to $tanh(c)$, where -1 is red and +1 is blue.}
\label{fig:sample}
\vspace{-0.15in}
\end{figure*}

\textbf{Interpretable, long-range LSTM cells.}
An LSTMs can in principle use its
memory cells to remember long-range information and keep track of various attributes of text it is currently processing. For instance, it
is a simple exercise to write down toy cell weights that would allow the cell to keep track of whether it is inside a
quoted string. However, to our knowledge, the existence of such cells has never been experimentally demonstrated
on real-world data. In particular, it could be argued that even if the LSTM is in principle capable of using these
operations, practical optimization challenges (i.e. SGD dynamics, or approximate gradients due to truncated backpropagation
through time) might prevent it from discovering these solutions. In this experiment we verify that multiple interpretable
cells do in fact exist in these networks (see Figure \ref{fig:sample}). For instance, one cell is clearly
acting as a line length counter, starting with a high value and then slowly decaying with each character until the next newline. Other
cells turn on inside quotes, the parenthesis after if statements, inside strings or comments, or with increasing strength as the
indentation of a block of code increases. In particular, note that truncated backpropagation with our hyperparameters
prevents the gradient signal from directly noticing dependencies longer than 100 characters, but we still observe
cells that reliably keep track of quotes or comment blocks much longer than 100 characters (e.g. $\sim 230$ characters in the
quote detection cell example in Figure \ref{fig:sample}). We hypothesize that these cells first develop on patterns shorter than
100 characters but then also appropriately generalize to longer sequences.

\begin{figure*}
\includegraphics[width=0.58\linewidth]{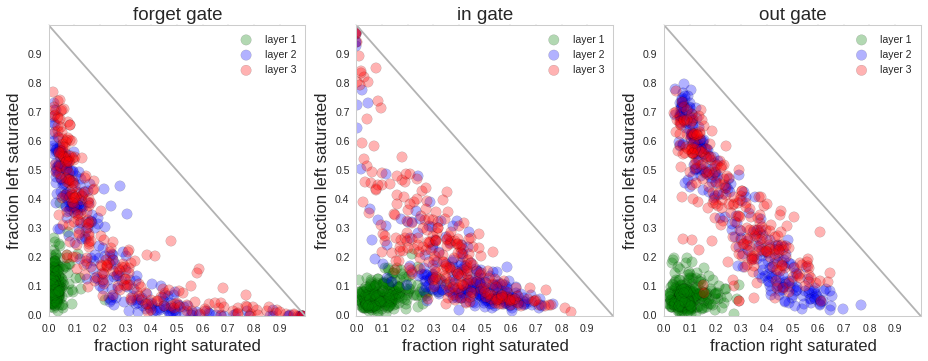}
\hspace{0.04\linewidth}
\includegraphics[width=0.38\linewidth]{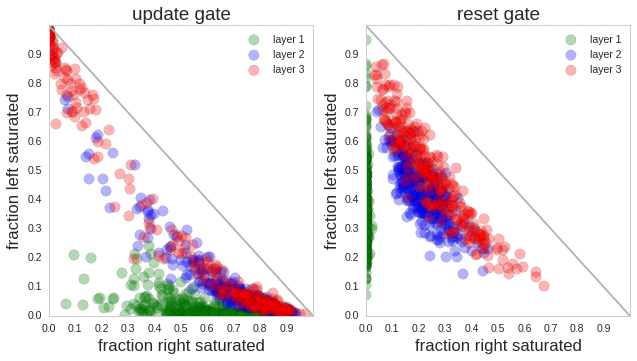}
\caption{
\textbf{Left three:} Saturation plots for an LSTM. Each circle is a gate in the LSTM and its position is
determined by the fraction of time it is left or right-saturated. These fractions must add to at most one
(indicated by the diagonal line). \textbf{Right two:} Saturation plot for a 3-layer GRU model.
}
\label{fig:saturations}
\vspace{-0.15in}
\end{figure*}

\textbf{Gate activation statistics}. We can gain some insight into the internal mechanisms of the LSTM by studying the gate
activations in the networks as they process test set data. We were particularly interested in
looking at the distributions of saturation regimes in the networks, where we define a gate to be
left or right-saturated if its activation is less than 0.1 or more than 0.9, respectively, or unsaturated otherwise. We then
compute the fraction of times that each LSTM gate spends left or right saturated, and plot the results in Figure \ref{fig:saturations}.
For instance, the number of often right-saturated forget gates is particularly interesting, since this corresponds to cells that
remember their values for very long time periods. Note that there are multiple cells that are almost always right-saturated
(showing up on bottom, right of the forget gate scatter plot), and hence function as nearly perfect integrators. Conversely,
there are no cells that function in purely feed-forward fashion, since their forget gates would show up as consistently left-saturated
(in top, left of the forget gate scatter plot). The output gate statistics also reveal that there are no cells that get consistently
revealed or blocked to the hidden state. Lastly, a surprising finding is that unlike the other two layers that contain gates with
nearly binary regime of operation (frequently either left or right saturated), the activations in the first layer are much more diffuse
(near the origin in our scatter plots). We struggle to explain this finding but note that it is present across all of our models. A similar
effect is present in our GRU model, where the first layer reset gates $r$ are nearly never right-saturated and the update gates $z$ are
rarely ever left-saturated. This points towards a purely feed-forward mode of operation on this layer, where the previous
hidden state is barely used.

\vspace{-0.1in}
\subsection{Understanding Long-Range Interactions}
\vspace{-0.1in}

Good performance of LSTMs is frequently attributed to their ability to store long-range information.
In this section we test this hypothesis by comparing an LSTM with baseline models that
can only utilize information from a fixed number of previous steps. In particular, we consider two baselines:

\hspace{0.1in} \textit{1. $n$-NN}: A fully-connected neural network with one hidden layer and $\mathrm{tanh}$
nonlinearities. The input to the network is a sparse binary vector of dimension
$nK$ that concatenates the one-of-$K$ encodings of $n$ consecutive characters. We optimize the model
as described in Section~\ref{subsec:optimization} and cross-validate the size of the hidden layer.

\hspace{0.1in} \textit{2. $n$-gram}: An unpruned $(n+1)$-gram language model
using modified Kneser-Ney smoothing \cite{chen1999empirical}. This is a standard smoothing
method for language models \cite{huang2001spoken}. All models were trained using the
popular KenLM software package \cite{Heafield-estimate}.

\setcounter{table}{1}
\begin{table*}[t]
\small
\centering
\begin{tabulary}{\linewidth}{L|CCCCCCCCC|C}
\backslashbox{Model}{$n$} & 1 & 2 & 3 & 4 & 5 & 6 & 7 & 8 & 9 & 20\\
\hline
\multicolumn{11}{c}{War and Peace Dataset} \\
\hline
$n$-gram & 2.399 & 1.928 & 1.521 & 1.314 & 1.232 & 1.203 & \textbf{1.194} & 1.194 & 1.194 & 1.195 \\
$n$-NN & 2.399 & 1.931 & 1.553 & 1.451 & 1.339 & \textbf{1.321} & - & - & - & - \\
\hline
\multicolumn{11}{c}{Linux Kernel Dataset} \\
\hline
$n$-gram & 2.702 & 1.954 & 1.440 & 1.213 & 1.097 & 1.027 & 0.982 & 0.953 & 0.933 & \textbf{0.889} \\
$n$-NN & 2.707 & 1.974 & 1.505 & 1.395 & \textbf{1.256} & 1.376 & - & - & - & - \\
\end{tabulary}
\caption{The \textbf{test set cross-entropy loss} on both datasets for $n$-gram models (low is good).
The standard deviation estimate using 100 bootstrap samples is below $4\times10^{-3}$ in all cases.}
\label{tab:ngram-performance}
\end{table*}

\textbf{Performance comparisons.}
The performance of both $n$-gram models is shown in Table~\ref{tab:ngram-performance}.
The $n$-gram and $n$-NN models perform nearly identically for small values of $n$,
but for larger values the $n$-NN models start to overfit and the $n$-gram model performs better.
Moreover, we see that on both datasets our best recurrent network outperforms the 20-gram
model (1.077 vs. 1.195 on WP and 0.84 vs.0.889). It is difficult to make a direct model size comparison,
but the 20-gram model file has 3GB, while our largest checkpoints are 11MB. However, the assumptions
encoded in the Kneser-Ney smoothing model are intended for word-level modeling of natural
language and may not be optimal for character-level data. Despite this concern, these results already provide
weak evidence that the recurrent networks are effectively utilizing information beyond 20 characters.

\begin{figure}
\centering
\includegraphics[width=0.19\linewidth]{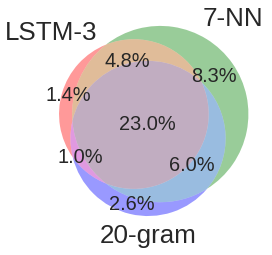}
\includegraphics[width=0.39\linewidth]{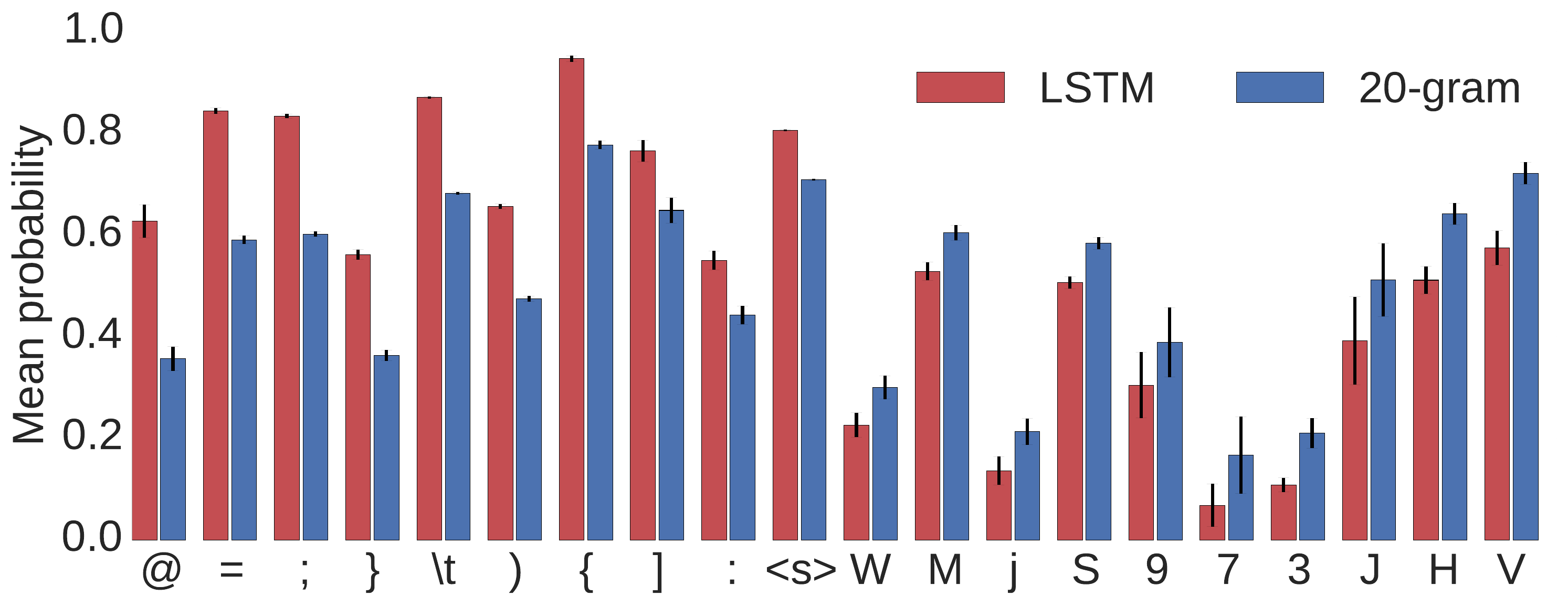}
\includegraphics[width=0.39\linewidth]{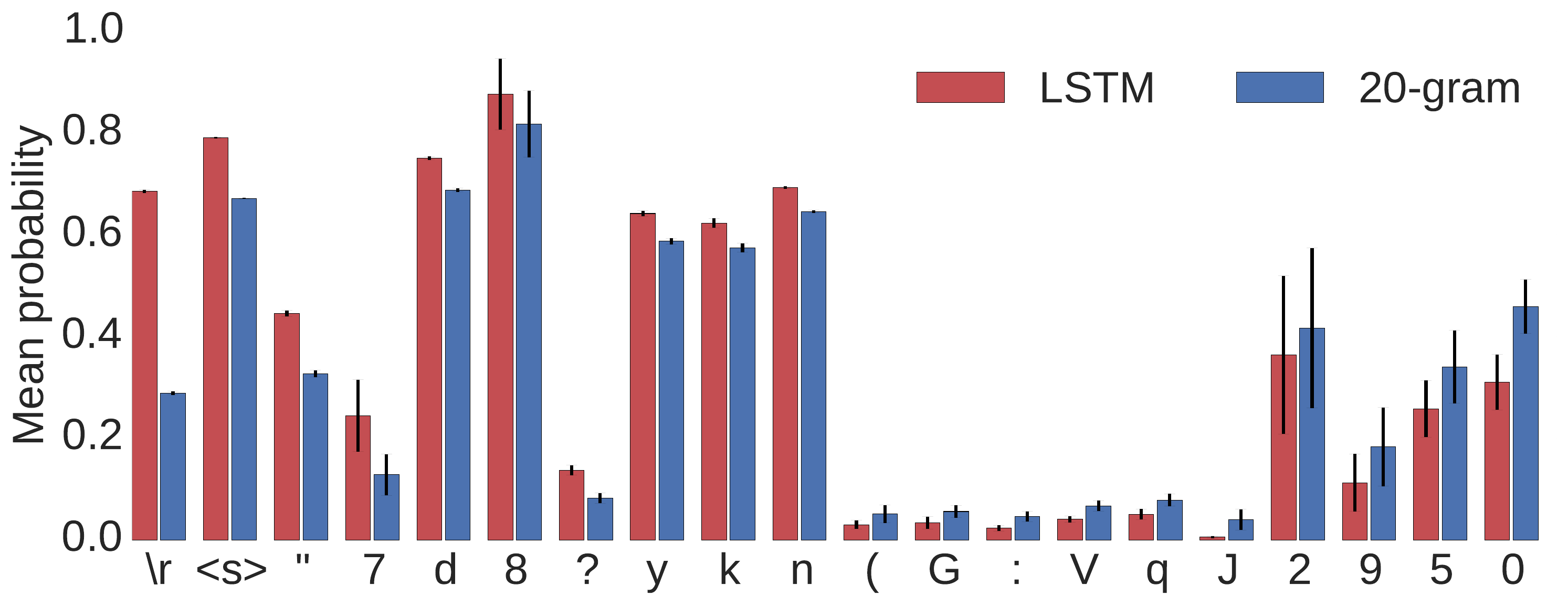}
\caption{
\textbf{Left:} Overlap between test-set errors between our best 3-layer LSTM and
the $n$-gram models (low area is good).
\textbf{Middle/Right:} Mean probabilities assigned to a correct character (higher is better),
broken down by the character, and then sorted by the difference between two models.
``\textless s\textgreater'' is the space character. LSTM (red) outperforms the
20-gram model (blue) on special characters that require long-range reasoning.
Middle: LK dataset, Right: WP dataset.
}
\label{fig:error-venn}
\vspace{-0.1in}
\end{figure}

\textbf{Error Analysis.}
It is instructive to delve deeper into the errors made by both recurrent networks and $n$-gram models.
In particular, we define a character to be an error if the probability assigned to it by a model on the previous time step
is below 0.5. Figure~\ref{fig:error-venn} (left) shows the overlap between the test-set errors for the
3-layer LSTM, and the best $n$-NN and $n$-gram models. We see that the majority of errors are
shared by all three models, but each model also has its own unique errors.

To gain deeper insight into the errors that are unique to the LSTM or the 20-gram model,
we compute the mean probability assigned to each character in the vocabulary across the test
set. In Figure~\ref{fig:error-venn} (middle,right) we display the 10 characters where each
model has the largest advantage over the other.
On the Linux Kernel dataset, the LSTM displays a large advantage on special characters that
are used to structure C programs, including whitespace and brackets. The War and Peace
dataset features an interesting long-term dependency with the carriage return, which occurs
approximately every 70 characters. Figure~\ref{fig:error-venn} (right) shows that the LSTM has a
distinct advantage on this character. To accurately predict the presence of the carriage return
the model likely needs to keep track of its distance since the last carriage return. The cell example
we've highlighted in Figure~\ref{fig:sample} (top, left) seems particularly well-tuned for
this specific task. Similarly, to predict a closing bracket or quotation mark, the model must be
aware of the corresponding open bracket, which may have appeared many time steps ago.
The fact that the LSTM performs significantly better than the 20-gram model on these characters
provides strong evidence that the model is capable of effectively keeping track of long-range interactions.

\begin{figure*}
\centering
\includegraphics[width=0.48\textwidth]{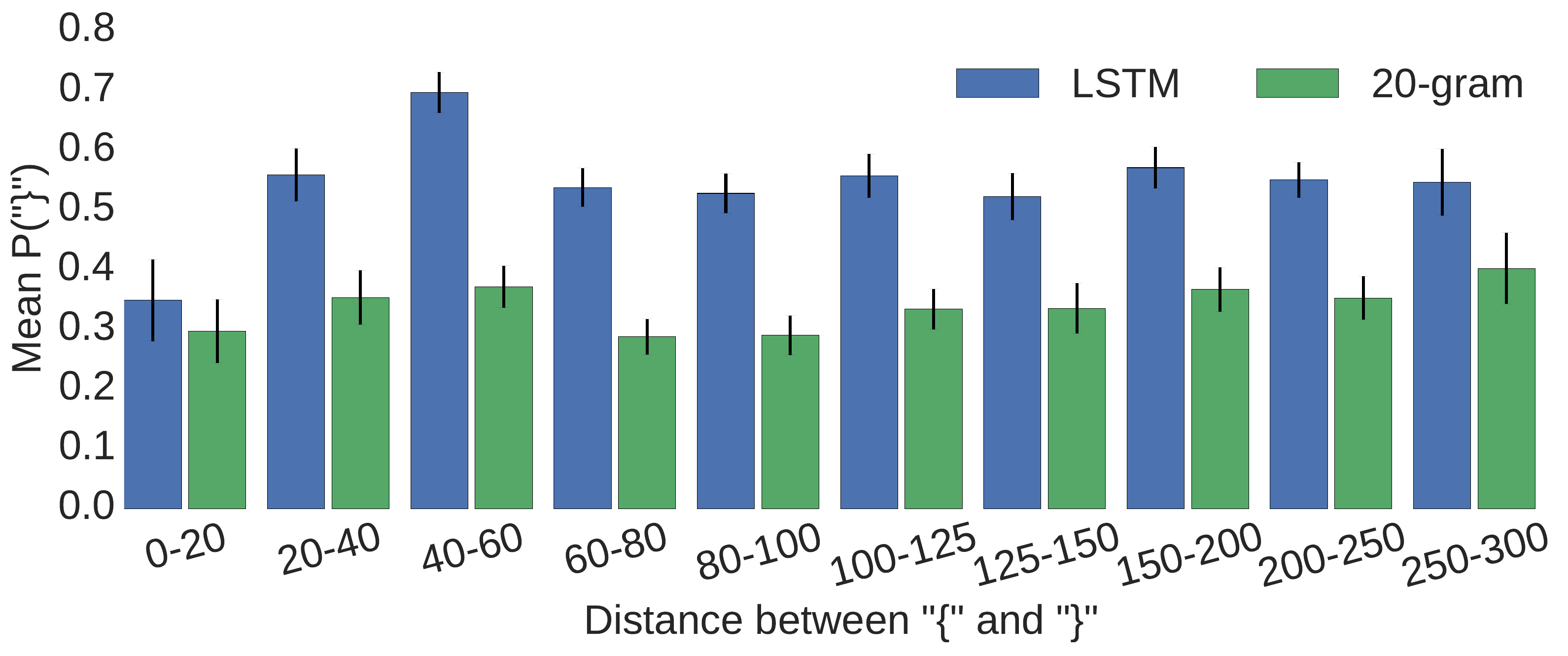}
\includegraphics[width=0.48\textwidth]{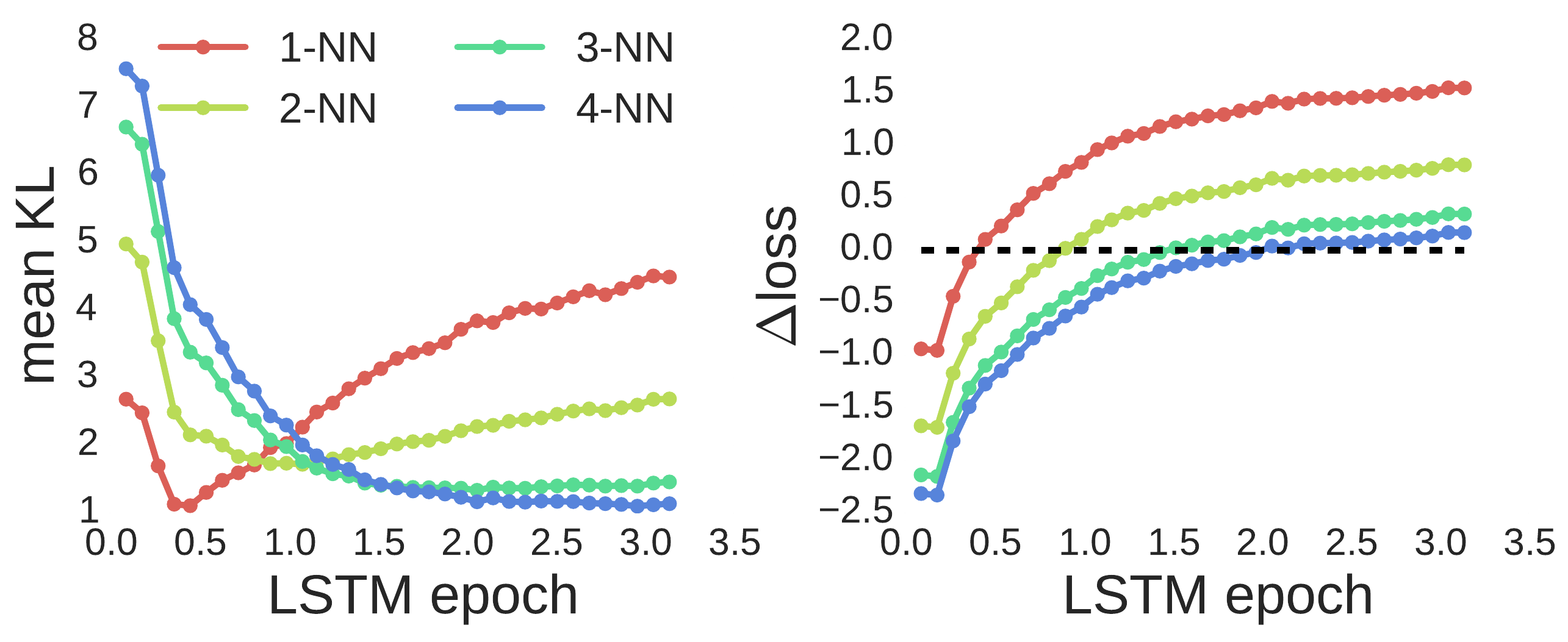}
\caption{
\textbf{Left}: Mean probabilities that the LSTM and 20-gram model assign to
the ``\}" character, bucketed by the distance to the matching ``\{".
\textbf{Right}: Comparison of the similarity between 3-layer LSTM and the $n$-NN baselines
over the first 3 epochs of training, as measured by the symmetric KL-divergence (middle) and the
test set loss (right). Low KL indicates similar predictions, and positive $\Delta$loss indicates that
the LSTM outperforms the baseline.
}
\label{fig:brace-distance}
\vspace{-0.15in}
\end{figure*}

\textbf{Case study: closing brace}. Of these structural characters,
the one that requires the longest-term reasoning is the closing
brace (``\}'') on the Linux Kernel dataset. Braces are used to denote blocks of code, and may be
nested; as such, the distance between an opening brace and its corresponding closing brace can
range from tens to hundreds of characters. This feature makes the closing brace an ideal test
case for studying the ability of the LSTM to reason over various time scales. We group closing
brace characters on the test set by the distance to their corresponding open brace and compute
the mean probability assigned by the LSTM and the 20-gram model to closing braces within each
group. The results are shown in Figure~\ref{fig:brace-distance} (left). First, note that the LSTM
only slightly outperforms the 20-gram model in the first bin, where the distance between braces
is only up to 20 characters. After this point the performance of the 20-gram model stays
relatively constant, reflecting a baseline probability of predicting the closing brace without seeing
its matching opening brace. Compared to this baseline, we see that the LSTM gains significant
boosts up to 60 characters, and then its performance delta slowly decays over time as it becomes
difficult to keep track of the dependence.

\textbf{Training dynamics.}
It is also instructive to examine the training dynamics of the LSTM by comparing it
with trained $n$-NN models during training using the (symmetric) KL divergence between the
predictive distributions on the test set. We plot the divergence and the difference in the
mean loss in Figure~\ref{fig:brace-distance} (right). Notably, we see that in the first few
iterations the LSTM behaves like the 1-NN model but then diverges from it soon after. The LSTM
then behaves most like the 2-NN, 3-NN, and 4-NN models in turn. This experiment suggests that
the LSTM ``grows'' its competence over increasingly longer dependencies during training.
This insight might be related to why Sutskever et al. \cite{sutskever2014sequence} observe
improvements when they reverse the source sentences in their encoder-decoder architecture for machine
translation. The inversion introduces short-term dependencies that the LSTM can model first,
and then longer dependencies are learned over time.

\vspace{-0.1in}
\subsection{Error Analysis: Breaking Down the Failure Cases}
\vspace{-0.1in}

In this section we break down LSTM's errors into categories to study the remaining limitations,
the relative severity of each error type, and to suggest areas for further study. We focus on
the War and Peace dataset where it is easier to categorize the errors. Our approach is to
\textit{``peel the onion''} by iteratively removing the errors with a series of constructed oracles.
As in the last section, we consider a character to be an error if the probability it was assigned
by the model in the previous time step is below 0.5. Note that the order in which the oracles are
applied influences the results. We tried to apply the oracles
in order of increasing difficulty of removing each error category and believe that the final results
are instructive despite this downside. The oracles we use are, in order:

\textbf{$n$-gram oracle.} First, we construct optimistic $n$-gram oracles that eliminate errors that
might be fixed with better modeling of short dependencies. In particular, we evaluate the $n$-gram
model ($n = 1, \ldots, 9$) and remove a character error if it is correctly classified (probability assigned
to that character greater than 0.5) by any of these models. This gives us an approximate idea of the
amount of signal present only in the last 9 characters, and how many errors we could optimistically
hope to eliminate without needing to reason over long time horizons.

\vspace{-0.07in}
\textbf{Dynamic $n$-long memory oracle.} To motivate the next oracle, consider the string \textit{``Jon yelled at Mary but Mary couldn't
hear him.''} One interesting and consistent failure mode that we noticed in the predictions is that if the LSTM fails to predict the characters
of the first occurrence of \textit{``Mary''} then it will almost always also fail to predict the same characters of the second occurrence,
with a nearly identical pattern of errors. However, in principle the presence of the first mention should make the second much
more likely. The LSTM could conceivably store a summary of previously seen characters in the data and fall back on this memory when it is
uncertain. However, this does not appear to take place in practice. This limitation is related to the improvements
seen in ``dynamic evaluation'' \cite{mikolov2012statistical,jelinek1991dynamic} of recurrent language models,
where an RNN is allowed to train on the test set characters during evaluation as long as it sees them
only once. In this mode of operation when the RNN trains on the first occurrence of \textit{``Mary''}, the log probabilities on the second
occurrence are significantly better. We hypothesize that this \textit{dynamic} aspect is a common feature of sequence data,
where certain subsequences that might not frequently occur in the training data should still be more likely if they
were present in the immediate history. However, this general algorithm does not seem to be learned by the LSTM. Our
dynamic memory oracle quantifies the severity of this limitation by removing errors in all words (starting with the second
character) that can found as a substring in the last $n$ characters (we use $n \in \{100, 500, 1000, 5000\})$.

\begin{figure*}
\centering
\includegraphics[width=1\linewidth]{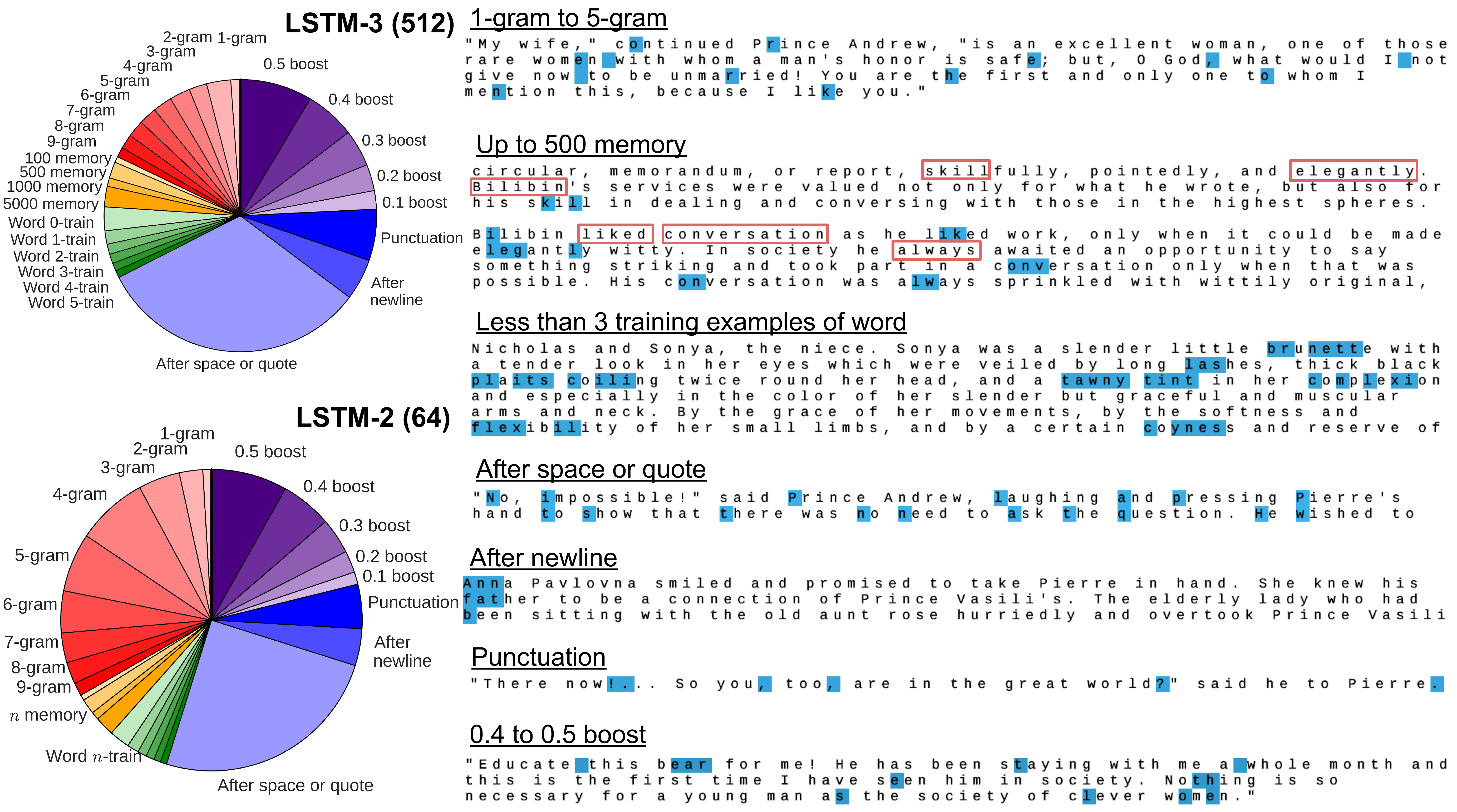}
\caption{
\textbf{Left:} LSTM errors removed one by one with oracles, starting from top of the pie chart
and going counter-clockwise. The area of each slice corresponds to fraction of errors contributed.
``$n$-memory'' refers to dynamic memory oracle with context of $n$ previous characters. ``Word $t$-train''
refers to the rare words oracle with word count threshold of $t$.
\textbf{Right:} Concrete examples of text from the test set for each error type. Blue color
highlights the relevant characters with the associated error. For the memory category we also
highlight the repeated substrings with red bounding rectangles.}
\label{fig:errors}
\vspace{-0.15in}
\end{figure*}

\vspace{-0.07in}
\textbf{Rare words oracle.} Next, we construct an oracle that eliminates errors for rare words
that occur only up to $n$ times in the training data ($n = 0, \ldots, 5$). This estimates the severity of errors
that could optimistically be eliminated by increasing the size of the training data, or with pretraining.

\vspace{-0.07in}
\textbf{Word model oracle.} We noticed that a large portion of the errors occur on the first character
of each word. Intuitively, the task of selecting the next word in the sequence is harder than completing
the last few characters of a known word. Motivated by this observation we constructed an oracle that
eliminated all errors after a space, quote or a newline. Interestingly, a high portion of errors can be
found after a newline, since the models have to learn that newline has semantics similar to a space.

\vspace{-0.07in}
\textbf{Punctuation oracle.} The remaining errors become difficult to blame on one particular, interpretable
aspect of the modeling. At this point we construct an oracle that removes errors on all punctuation.

\vspace{-0.07in}
\textbf{Boost oracles.} The remaining errors that do not show salient structures or patterns are removed by an
oracle that boosts the probability of the correct letter by a fixed amount.
These oracles allow us to understand the distribution of the difficulty of the remaining errors.

We now subject two LSTM models to the error analysis: First, our best LSTM model and second, the best LSTM model
in the smallest model category (50K parameters). The small and large models allow us to understand how the error
break down changes as we scale up the model. The error breakdown after applying each oracle for both models
can be found in Figure \ref{fig:errors}.

\textbf{The error breakdown.} In total, our best LSTM model made a total of 140K
errors out of 330K test set characters (42\%). Of these, the $n$-gram oracle eliminates 18\%, suggesting that the model is not
taking full advantage of the last 9 characters. The dynamic memory oracle eliminates 6\% of the errors. In principle,
a dynamic evaluation scheme could be used to mitigate this error, but we believe that a more principled approach
could involve an approach similar to Memory Networks \cite{memorynets}, where the model is allowed to attend to a recent
history of the sequence while making its next prediction. Finally, the rare words oracle accounts for 9\% of the errors.
This error type might be mitigated with unsupervised pretraining \cite{dai2015semi}, or by increasing the size of the training set.
The majority of the remaining errors (37\%) follow a space, a quote, or a newline, indicating the model's difficulty
with word-level predictions. This suggests that longer time horizons in backpropagation through time, or possibly hierarchical
context models, could provide improvements. See Figure \ref{fig:errors} (right) for examples of each error type. We
believe that this type of error breakdown is a valuable tool for isolating and understanding the source of improvements
provided by new proposed models.

\textbf{Errors eliminated by scaling up}. In contrast, the smaller LSTM
model makes a total of 184K errors (or 56\% of the test set), or approximately 44K more than the large model.
Surprisingly, 36K of these errors (81\%) are $n$-gram errors, 5K come from the boost category,
and the remaining 3K are distributed across the other categories relatively evenly.
That is, scaling the model up by a factor 26 in the number of parameters has almost entirely provided gains
in the local, $n$-gram error rate and has left the other error categories untouched in comparison.
This analysis provides some evidence that it might be necessary to develop new architectural improvements
instead of simply scaling up the basic model.

\vspace{-0.1in}
\section{Conclusion}
\vspace{-0.1in}

We have used character-level language models as an interpretable test bed for analyzing the predictions, representations
training dynamics, and error types present in Recurrent Neural Networks. In particular, our qualitative visualization
experiments, cell activation statistics and comparisons to finite horizon $n$-gram models demonstrate that these
networks learn powerful, and often interpretable long-range interactions on real-world data. Our error analysis broke
down cross entropy loss into several interpretable categories, and allowed us to illuminate the sources of remaining
limitations and to suggest further areas for study. In particular, we found that scaling up the model almost entirely
eliminates errors in the $n$-gram category, which provides some evidence that further architectural innovations may 
be needed to address the remaining errors. 

\subsubsection*{Acknowledgments}

We gratefully acknowledge the support of NVIDIA Corporation with the donation of the GPUs used for this research.

\bibliography{iclr2016_conference}
\bibliographystyle{iclr2016_conference}

\end{document}